\documentclass[11pt]{article}

\usepackage[final]{acl}

\usepackage{times}
\usepackage{latexsym}

\usepackage{hyperref}

\usepackage[T1]{fontenc}

\usepackage[utf8]{inputenc}

\usepackage{microtype}

\usepackage{inconsolata}

\usepackage{graphicx}

\usepackage{multirow}

\usepackage[table]{xcolor}

\title{HG-RAG: Hierarchy-Guided Retrieval-Augmented Generation for Structured Knowledge Graphs}

\author{Pranav Yadav \\
  University of California, Merced \\
  Merced, CA, USA \\
  \texttt{pyadav@ucmerced.edu}
}

\begin{document}
\maketitle
\begin{abstract}
Retrieval Augmented Generation (RAG) has proven to be a widely successful process at improving the quality of outputs from a Large Language Model (LLM) for wider context. However, RAG systems typically retrieve context from flat document stores, which struggles when queries require hierarchical or relational reasoning across structured knowledge. I present HG-RAG (Hierarchy-Guided RAG), a framework that performs graph-traversal over a hierarchical knowledge graph to deliver structured context to a language model. My retrieval pipeline resolves a named entity anchor from the query, then expands context upward through parent nodes, laterally through relational neighbors, and downward through child nodes when needed. I evaluate HG-RAG against a dense retrieval baseline across three world scales (18–800 nodes) with four query types:  \texttt{local fact},  \texttt{hierarchical},  \texttt{neighborhood}, and  \texttt{multi\_hop}. Results show HG-RAG consistently outperforms the flat baseline on hierarchical, relational, and multi-hop reasoning tasks, while reducing hallucination and maintaining locality coherence.
\end{abstract}

\url{https://github.com/Pranubot/HG-RAG}

\section{Introduction}

LLMs can often provide incorrect information when presented with a query since their knowledge consists purely of pre-training data \citep{ji-etal-2023-hallucination, shuster2021retrievalaugmentationreduceshallucination}. A common workaround is to provide LLMs with additional context upfront with the query. However, the effectiveness of this method starts to break down when the context is too large to be tacked on the side, requiring a smarter system. Retrieval Augmented Generation aims to reduce this context build-up by using a user’s query to search for semantically similar data, which are added to the model’s context. This avoids the whole context, including data irrelevant to the query, from being attached.

Existing RAG frameworks assume flat, unstructured retrieval. When knowledge is inherently hierarchical (geopolitical structures, taxonomies, tree structures), dumping a flat context window onto your LLM loses relational signal and increases risk of hallucination  \citep{shuster2021retrievalaugmentationreduceshallucination}. Structured knowledge demands structured retrieval; a system that understands not just what a node is, but where it sits in the hierarchy. I address this by extending the RAG framework with graph-traversal subroutines that collect context from a named entity anchor upward through parent nodes, laterally through relational neighbors, and downward through children. This produces a focused subgraph of contextually relevant nodes rather than an unorganized document dump, giving the LLM the relational scaffolding it needs to reason accurately across hierarchical and multi-hop queries (Multi-hop queries require chaining multiple relational steps across the graph to arrive at an answer).

\begin{figure}[h]
  \centering
  \includegraphics[width=1.2\columnwidth]{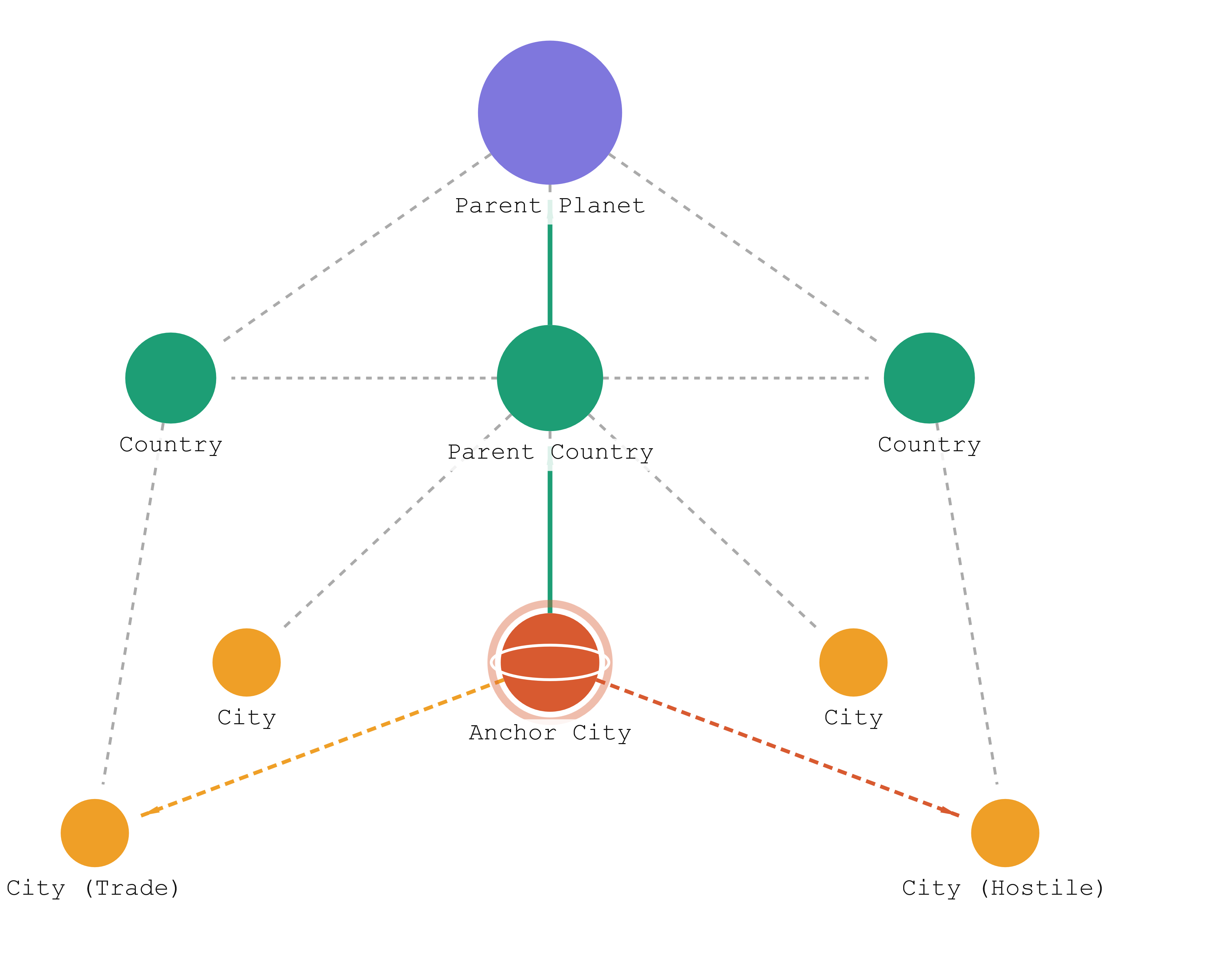}
  \caption{A small snapshot of what kind of graphs we are retrieving from and how HG-RAG visualizes them. We visualize relationships relative to our anchor node (red).}
  \label{fig:your-label}
\end{figure}

\section{Related Work}

Retrieval-Augmented Generation (RAG) was introduced by \citet{lewis-etal-2020-rag} as a method for grounding LLM responses in external knowledge by retrieving relevant documents at inference time. While effective for open-domain question answering, standard RAG requires retrieved documents to be flat, unordered context. 

Several works have explored extending RAG to graph-structured knowledge. Microsoft's GraphRAG \cite{edge-etal-2024-graphrag} constructs a community-based knowledge graph from source documents and uses it to guide retrieval. This proved to be a fruitful development on the standard naive RAG framework. Similarly, knowledge graph question answering systems such as KGQA \citep{lan-jiang-2020-query} have shown that explicit graph traversal improves multi-hop reasoning by following relational edges. In a related direction, \citet{yasunaga-etal-2021-qa} jointly trains a graph neural network alongside a language model to reason over knowledge graph subgraphs. In contrast, dense retrieval methods \citep{karpukhin-etal-2020-dpr} rely on semantic similarity in a shared vector space, with no mechanism for exploiting relational structure. A broader survey of neural KGQA approaches \citep{chakraborty-etal-2021-kgqa} further demonstrates that graph-structured retrieval consistently outperforms flat retrieval on relational reasoning tasks.

HG-RAG differs from these approaches in its explicit focus on \textit{hierarchical} structure. Rather than treating all graph edges as equivalent, HG-RAG distinguishes between structural edges (\texttt{contains}) and relational edges (\texttt{borders}, \texttt{trade\_with}, \texttt{hostile}), and uses a directional k-step traversal to collect context that respects the hierarchy. This makes HG-RAG exceptional for domains where knowledge is naturally organized into parent-child levels.

\section{Setup}

\subsection{World Construction}
To evaluate HG-RAG in a controlled setting, I construct a series of synthetic hierarchical knowledge graphs using a three-tier structure: Planets → Countries → Cities. These worlds are implemented as directed graphs using \texttt{NetworkX} \citep{hagberg2008networkx}. This hierarchy mirrors real-world geopolitical relations and provides a proper testbed for hierarchical reasoning. The framing is inspired by grand strategy video games where dense relational data between territories on multiple levels is the norm, making it a strong stress test for structured retrieval. 

Each node carries domain-relevant attributes that the LLM will be assessed on. Planets store population, stability, and exports; cities store imports. Edges include both structural and relational information of six types: \texttt{contains} (children), \texttt{borders}, \texttt{trade\_with}, \texttt{hostile}, \texttt{unfriendly}, and \texttt{neutral}. To guarantee valid ground-truth answers for multi-hop queries, city imports are post-assigned from their trade partners’ exports after world generation. This is a semantic coupling heuristic that ensures every trade-chain query has a deterministic, resolvable answer grounded in the graph.

\begin{table}[h!]
  \centering
  \small
  \begin{tabular}{lcccc}
    \hline
    \textbf{Size} & \textbf{Planets} & \textbf{Countries} & \textbf{Cities} & \textbf{Total Nodes} \\
    \hline
    Small  & 2 & 3 & 3  & $\sim$18  \\
    \hline
    Medium & 3 & 5 & 10 & $\sim$150 \\
    \hline
    Large  & 4 & 8 & 25 & $\sim$800 \\
    \hline
  \end{tabular}
  \caption{World graph sizes used in testing.}
  \label{tab:kg-sizes}
\end{table}

This range allows us to observe how both retrieval precision and entity resolution change with scale. 

\subsection{LLM Model}
\begin{table*}[t]
  \centering
  \setlength{\tabcolsep}{20pt}
  \begin{tabular}{lp{5cm}l}
    \hline
    \textbf{Type} & \textbf{Example} & \textbf{Anchor} \\
    \hline
    \texttt{local\_fact}    & ``What does \{city\} export?''                          & city \\
    \hline
    \texttt{hierarchical}   & ``What country is \{city\} in?''                        & city \\
    \hline
    \texttt{neighborhood}   & ``Which cities trade with \{city\}?''                   & city \\
    \hline
    \texttt{multi\_hop}     & ``A strike halts \{exports\} ... which cities are affected?'' & city or country \\
    \hline
  \end{tabular}
  \caption{Query types used in testing.}
  \label{tab:query-types}
\end{table*}

The open-source LLM model I have chosen for both our Baseline and RAG systems is Ollama's \texttt{Mistral 7B}, a high-performance 7.3 billion parameter language model. I chose this model due to its recency and high performance given the smaller size. For my purposes, it was crucial to select a model that wasn't over-engineered for the desired
task or too large for the average enthusiast to download for reproducing results. 

\subsection{RAG Retrieval Pipeline}
HG-RAG operates in 4 stages when given a natural language query:

\begin{enumerate}
    \item \textbf{Entity Resolution.} An LLM call scans the query for a named entity, which becomes the \textit{anchor node} for traversal. If the LLM fails to extract a match, a fallback chain is used: exact match → first-word match →  \texttt{difflib} fuzzy match (\texttt{cutoff=0.5}) → No entity found.  If the fuzzy match returns a node of the wrong entity type (ex: a city when a country is expected), the system silently falls back to the known ground-truth anchor. This is particularly important in large-world settings where name collisions are more likely, but also a limitation of the system worth mentioning.
    \item \textbf{Hierarchy Anchoring (\texttt{k-up}).} After an anchor node is established, the pipeline traverses upwards through “contains” edges, collecting parent and grandparent nodes up to k steps. My default for k-steps upwards is 2, this captures the full three-level 
    hierarchy.
    \item \textbf{Lateral Traversal (\texttt{k-side}).} The process expands horizontally, collecting relational neighbors up to k steps along \texttt{borders}, \texttt{trade-with}, \texttt{hostile}, \texttt{unfriendly}, and \texttt{neutral} edges. To prevent adversarial context, which is the most important information for our queries, from being diluted by neutral neighbors when the subgraph cap is reached, hostile and unfriendly neighbors are always prioritized for inclusion. 
    \item \textbf{Child Traversal (\texttt{k\_down}).} The pipeline can optionally collect child nodes downward through “contains” edges up to “k-down” steps. When the query requests such information, \texttt{k-down} increases from 0, which is the default. 
\end{enumerate}
The resulting subgraph is capped at 15 nodes to prevent context bloat. Retrieved nodes are serialized into a structured prompt block containing an explicit location chain (\texttt{"CityX is a city located in CountryY, on Planet Z"}), followed by labeled \texttt{[PLANET]}, \texttt{[COUNTRY]}, \texttt{[CITY]}, and \texttt{[RELATIONS]} blocks. In testing, arrow notion proved insufficient for LLM comprehension, prompting me to develop a different serialization.

\subsection{Baseline}

We compare HG-RAG against a dense vector RAG baseline that operates over the same knowledge graph without any structural awareness. At index time, each graph node (city, country, or planet) is serialized into a flat text chunk containing its attributes and direct relations, then encoded into a dense vector using \texttt{nomic-embed-text} \cite{nussbaum2024nomic} via Ollama. At query time, the question is embedded with the same model and cosine similarity is used to retrieve the top-10 most similar chunks, which are concatenated and passed as context to the LLM. The value $k{=}10$ is chosen to match the approximate context size of HG-RAG's subgraph retrieval for a fair comparison. Unlike HG-RAG, the baseline has no knowledge of the hierarchy: it cannot walk up from a city to its country or planet, cannot prioritize structurally critical relationships, and relies entirely on embedding similarity to surface relevant nodes.

\begin{figure*}[t]
\centering
\includegraphics[width=\textwidth]{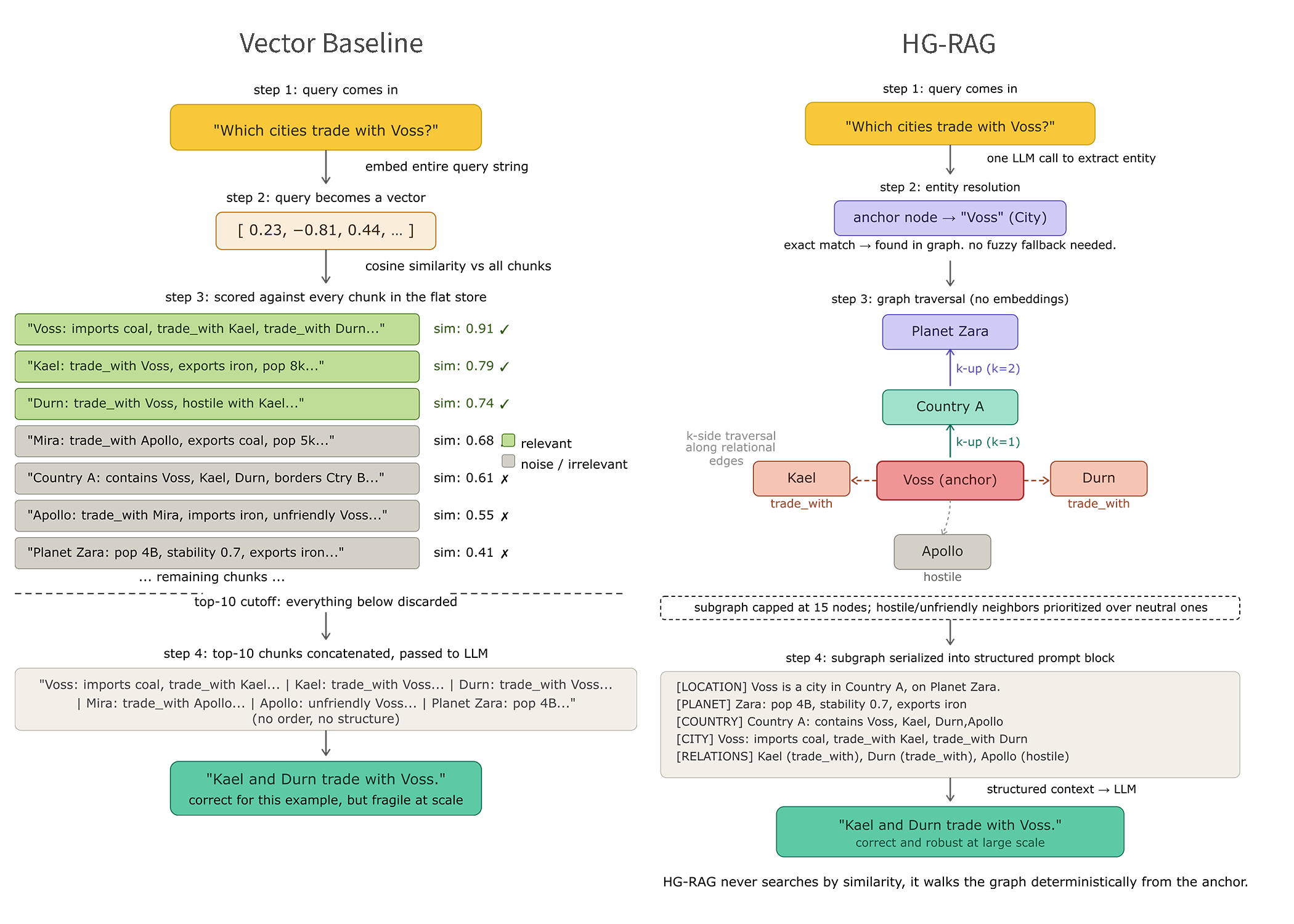}
\caption{Side-by-side comparison of the vector baseline and HG-RAG retrieval pipelines for the query ``Which cities trade with Voss?'' The baseline embeds the query, ranks all chunks by cosine similarity, and passes the top-10 as flat context. HG-RAG resolves the named entity, traverses the hierarchy upward through \texttt{contains} edges, and expands laterally through relational edges, producing a structured prompt with explicit location chains and relevant labels.}
\label{fig:pipeline-comparison}
\end{figure*}

\begin{table*}[t]
  \centering
  \small
  \begin{tabular}{llccccc}
    \hline
    \textbf{World Size} & \textbf{Query Type} & \textbf{System} & \textbf{Factual Accuracy} & \textbf{Hallucination Rate} & \textbf{Locality Awareness} \\
    \hline
    \multirow{10}{*}{Small}
      & \multirow{2}{*}{Hierarchical} & \cellcolor{gray!15}Baseline & \cellcolor{gray!15}0.50 & \cellcolor{gray!15}0.000 & \cellcolor{gray!15}1.00 \\
      &                               & HG-RAG   & 2.00 & 0.000 & 1.00 \\
      & \multirow{2}{*}{Local Fact}   & \cellcolor{gray!15}Baseline & \cellcolor{gray!15}1.00 & \cellcolor{gray!15}0.021 & \cellcolor{gray!15}1.00 \\
      &                               & HG-RAG   & 2.00 & 0.042 & 1.00 \\
      & \multirow{2}{*}{Multi-Hop}    & \cellcolor{gray!15}Baseline & \cellcolor{gray!15}1.18 & \cellcolor{gray!15}0.053 & \cellcolor{gray!15}0.88 \\
      &                               & HG-RAG   & 1.42 & 0.030 & 1.00 \\
      & \multirow{2}{*}{Neighborhood} & \cellcolor{gray!15}Baseline & \cellcolor{gray!15}1.33 & \cellcolor{gray!15}0.000 & \cellcolor{gray!15}1.00 \\
      &                               & HG-RAG   & 1.75 & 0.000 & 1.00 \\
      & \multirow{2}{*}{\textbf{Overall}} & \cellcolor{gray!15}Baseline & \cellcolor{gray!15}1.00 & \cellcolor{gray!15}0.018 & \cellcolor{gray!15}0.97 \\
      &                               & HG-RAG   & 1.79 & 0.018 & 1.00 \\
    \hline
    \multirow{10}{*}{Medium}
      & \multirow{2}{*}{Hierarchical} & \cellcolor{gray!15}Baseline & \cellcolor{gray!15}0.00 & \cellcolor{gray!15}0.000 & \cellcolor{gray!15}1.00 \\
      &                               & HG-RAG   & 2.00 & 0.000 & 1.00 \\
      & \multirow{2}{*}{Local Fact}   & \cellcolor{gray!15}Baseline & \cellcolor{gray!15}0.00 & \cellcolor{gray!15}0.000 & \cellcolor{gray!15}1.00 \\
      &                               & HG-RAG   & 2.00 & 0.125 & 1.00 \\
      & \multirow{2}{*}{Multi-Hop}    & \cellcolor{gray!15}Baseline & \cellcolor{gray!15}0.55 & \cellcolor{gray!15}0.072 & \cellcolor{gray!15}0.74 \\
      &                               & HG-RAG   & 1.58 & 0.035 & 1.00 \\
      & \multirow{2}{*}{Neighborhood} & \cellcolor{gray!15}Baseline & \cellcolor{gray!15}0.42 & \cellcolor{gray!15}0.008 & \cellcolor{gray!15}0.97 \\
      &                               & HG-RAG   & 1.58 & 0.000 & 1.00 \\
      & \multirow{2}{*}{\textbf{Overall}} & \cellcolor{gray!15}Baseline & \cellcolor{gray!15}0.24 & \cellcolor{gray!15}0.020 & \cellcolor{gray!15}0.93 \\
      &                               & HG-RAG   & 1.79 & 0.040 & 1.00 \\
    \hline
    \multirow{10}{*}{Large}
      & \multirow{2}{*}{Hierarchical} & \cellcolor{gray!15}Baseline & \cellcolor{gray!15}0.00 & \cellcolor{gray!15}0.000 & \cellcolor{gray!15}1.00 \\
      &                               & HG-RAG   & 2.00 & 0.000 & 1.00 \\
      & \multirow{2}{*}{Local Fact}   & \cellcolor{gray!15}Baseline & \cellcolor{gray!15}0.00 & \cellcolor{gray!15}0.000 & \cellcolor{gray!15}1.00 \\
      &                               & HG-RAG   & 2.00 & 0.042 & 1.00 \\
      & \multirow{2}{*}{Multi-Hop}    & \cellcolor{gray!15}Baseline & \cellcolor{gray!15}0.10 & \cellcolor{gray!15}0.025 & \cellcolor{gray!15}0.75 \\
      &                               & HG-RAG   & 1.50 & 0.029 & 1.00 \\
      & \multirow{2}{*}{Neighborhood} & \cellcolor{gray!15}Baseline & \cellcolor{gray!15}0.00 & \cellcolor{gray!15}0.000 & \cellcolor{gray!15}1.00 \\
      &                               & HG-RAG   & 1.87 & 0.000 & 1.00 \\
      & \multirow{2}{*}{\textbf{Overall}} & \cellcolor{gray!15}Baseline & \cellcolor{gray!15}0.02 & \cellcolor{gray!15}0.006 & \cellcolor{gray!15}0.94 \\
      &                               & HG-RAG   & 1.86 & 0.017 & 1.00 \\
    \hline
  \end{tabular}
  \caption{Average performance across 5 trials by query type and world size. Factual Accuracy (0--2), Hallucination Rate (0--1), Locality Awareness (0--1). Overall rows are bolded.}
  \label{tab:results}
\end{table*}

\begin{table*}[t]
  \centering
  \small
  \begin{tabular}{llccccc}
    \hline
    \textbf{World Size} & \textbf{Query Type} & \textbf{System} & \textbf{Small} & \textbf{Medium} & \textbf{Large} & \textbf{Average} \\
    \hline
    \multirow{2}{*}{Multi-Hop} & \multirow{2}{*}{LLM Judge (1--5)} & Baseline & 3.45 & 2.82 & 1.66 & 2.64 \\
                               &                                    & HG-RAG   & 3.23 & 3.88 & 4.10 & 3.74 \\
    \hline
  \end{tabular}
  \caption{Multi-hop LLM judge scores (1--5) across world sizes.}
  \label{tab:multihop}
\end{table*}

\section{Evaluation Metrics}

\subsection{Query Design}
We evaluate HG-RAG across four query types balanced at 25\% each of the total 50 question pool per world size. Each query type tackles a specific challenge regarding traversing a hierarchical graph. Furthermore, all queries are deterministically generated from the graph at construction time, ensuring the benchmark is fully reproducible and that every query has a verifiable ground-truth answer.

\texttt{local\_fact}: Local fact queries test whether the LLM can retrieve a single attribute directly from the anchor node. This is the most basic query type out of the 4, and does not require any context outside of the anchor.

\texttt{hierarchical}: Hierarchical queries require the system to traverse upward through the \texttt{contains} edge to identify a parent node. 

\texttt{neighborhood}: Neighborhood queries test lateral local traversal, requiring the LLM to identify relational neighbors along \texttt{trade\_with} or \texttt{borders} edges. 

\texttt{multi\_hop}: Multi-hop queries are the most demanding type, requiring the LLM to chain multiple relational steps: resolving an export, finding relevant trade partners, then identifying dependent cities. 

To prevent incoherent queries on smaller worlds, multi-hop queries are only generated when their preconditions are satisfied: a trade disruption query requires the anchor's export to be imported by at least one trade partner, and a war query requires at least one hostile or unfriendly neighbor. Small worlds would regularly have scenarios where there were no suitable entities to fulfill the pre-made prompts for multi-hop queries, so if not possible the program would fall back on any of the three other types of queries. Each multi-hop query is accompanied by a deterministic answer key built directly from graph data at generation time, which is passed to the LLM judge during evaluation.

\subsection{Rubric}
I evaluate LLM performance through four metrics, all tackling different dimensions of quality:

\begin{enumerate}

\item \textbf{Factual accuracy} (\texttt{factual\_accuracy}, 0-2) measures keyword overlap between the model’s response and ground-truth answer. A full score of 2 is awarded when overlap reaches $\geq 70\%$, 1 for $\geq 30\%$, and 0 for anything below. This metric captures whether the model includes the correct entities and values from the retrieved context

\item \textbf{Hallucination Rate} (\texttt{hallucination\_rate}, 0-1) measures the fraction of capitalized entity-like tokens in the response that do not appear anywhere in the knowledge graph. The lower the score, the better. This penalizes the model for inventing plausible-sounding entities that simply do not exist, a common failure when flat context overwhelms the LLM with loosely relevant information \cite{ji-etal-2023-hallucination}.

\item \textbf{Locality Awareness} (\texttt{locality\_awareness}, 0-1) measures the fraction of graph nodes mentioned in the response that belong to the anchor’s planet. This captures whether the model stays grounded in the relevant region of the hierarchy rather than drifting to unrelated parts of the world. I included this metric since it's a particularly important signal in large-world settings where many similar entities exist across different planets.

\item \textbf{LLM Judge Score} (\texttt{llm\_judge\_score}, 1-5) is utilized exclusively for multi-hop queries, where keyword overlap alone is insufficient to assess reasoning quality. Due to the open-ended nature of multi-hop questions, I created a special metric for them. The judge model receives the query, the model's response, and a deterministic answer key; then scores the response on a 1–5 rubric: 5 for correct entities with sound reasoning, 1 for wrong entities without reasoning. The rubric deliberately weights entity identification over reasoning quality, since LLMs are known to fabricate plausible reasoning chains even when their factual grounding is incorrect.

\end{enumerate}

One limitation worth noting: I use the same \texttt{Mistral 7B} instance as both the answering model and the LLM judge. While this is a practical constraint, it introduces the possibility of self-consistency bias \citep{zheng-etal-2023-judging}: the judge may be more lenient toward responses that mirror its own reasoning patterns. I treat \texttt{llm\_judge\_score} as a supplementary signal alongside the deterministic metrics rather than a primary measure of correctness.

\section{Experiment and Results}
Five trials were conducted per world size (small, medium, large), with 50 deterministically generated queries per trial split equally across the four query types. All experiments were run on a \texttt{2024 ASUS ROG Zephyrus G14} (\texttt{AMD Ryzen 9 8945HS}, \texttt{NVIDIA RTX 4060 Laptop GPU}, \texttt{16GB RAM}), with \texttt{Mistral 7B} operated locally via \texttt{Ollama} \citep{ollama2023} at \texttt{Q4\_K\_M}. Runtime for one trial was approximately 5 minutes.

\subsection{Baseline Collapse at Scale}
The embedding-based RAG baseline performs reasonably on small worlds, achieving an overall factual accuracy of 1.004. However, this performance rapidly deteriorates with scale: factual accuracy falls to 0.242 on medium worlds and to 0.022 on large worlds. Despite using semantic retrieval to surface relevant chunks, the baseline scores 0.00 on hierarchical and local fact queries at both medium and large scales. This suggests that cosine similarity over flat node descriptions fails to consistently retrieve structurally relevant nodes; for example, the parent country for a city may not be semantically close to the query, even though it is the correct answer. The baseline's strongest results occur on multi-hop queries, where it achieves 1.184 on small worlds, but even here performance degrades to 0.100 at large scale.

One notable pattern in the baseline's failure mode is that hallucination rates remain low even as factual accuracy collapses. At large scale, the baseline achieves a hallucination rate of just 0.006 while scoring near-zero on accuracy. This suggests that when semantic retrieval surfaces the wrong nodes, the LLM faithfully reports their contents rather than inventing entities, producing answers that are internally consistent but factually incorrect. This suggests that my hallucination rate metric alone is not a reliable proxy for answer quality.

\subsection{HG-RAG Performance}
HG-RAG maintains strong performance across all world sizes. Overall factual accuracy is 1.792 for small worlds, 1.792 for medium, and 1.857 for large. On hierarchical and local fact queries, HG-RAG achieves a perfect 2.00 at every scale, confirming that the traversal pipeline reliably surfaces the anchor node and its structural parents. Neighborhood queries show consistent gains as well, rising from 1.750 at small scale to 1.867 at large scale.
 
Locality awareness remains at a perfect 1.00 across all world sizes and query types for HG-RAG, meaning the system never drifts to entities outside the anchor's planet. The baseline, by contrast, shows locality degradation on multi-hop queries in particular, dropping to 0.75 at large scale. This confirms that hierarchical anchoring keeps the LLM grounded in the relevant region of the graph.
 
Hallucination rates for HG-RAG are generally low, averaging 0.018 on small worlds, 0.040 on medium, and 0.017 on large. One outlier is the local fact hallucination rate of 0.125 on medium worlds, which is higher than any other HG-RAG condition. This warrants further investigation through further research, but does not undermine the broader trend.

\subsection{Multi-Hop Scaling}
Multi-hop queries have proved to be the most demanding type for both systems, requiring the LLM to chain multiple relational steps to arrive at an answer. The LLM judge scores reveal an inverse scaling pattern between the two systems. The baseline degrades as world size increases (3.450 $\rightarrow$ 2.817 $\rightarrow$ 1.660), consistent with its broader collapse under increased noise. HG-RAG, by contrast, improves with scale (3.233 $\rightarrow$ 3.883 $\rightarrow$ 4.100), achieving its strongest multi-hop performance on the largest and most complex world. 

\subsection{Why Small Worlds Narrow the Gap}
 
An initially counterintuitive finding is that HG-RAG does not perform notably better on small worlds than on medium or large ones. You may expect a smaller graph would make retrieval easier due to less noise. Two factors contribute to this. First, with only $\sim$18 nodes, the 15-node subgraph cap means HG-RAG retrieves nearly the entire graph, eroding the signal-to-noise advantage that hierarchical traversal provides at larger scales. The retrieved context, despite being structured and locally anchored, contains a high proportion of nodes only loosely relevant to the query. Second, the baseline itself faces less noise in small worlds. With fewer entities competing for attention, semantic retrieval is less likely to surface irrelevant chunks. In larger worlds, both effects reverse: the subgraph cap acts as a meaningful filter that isolates a tight neighborhood from a vast graph, while the baseline's retrieval struggles to identify structurally relevant nodes among hundreds of candidates. This explains why HG-RAG's strongest results occur at large scale rather than small scale: scale amplifies the benefit of structured retrieval rather than undermining it.

\section{Conclusion}
This paper presented HG-RAG, a retrieval-augmented generation framework that applies graph-traversal heuristics over a hierarchical knowledge graph to deliver structured and relevant context to a language model. When evaluated across three world scales and four query types, HG-RAG consistently outperformed an embedding-based RAG baseline that retrieves context via cosine similarity over dense node vectors. While the baseline achieves reasonable performance on small worlds, its factual accuracy collapses to 0.022 at large scale, even with semantic retrieval actively selecting the top-k most relevant chunks. Conversely, HG-RAG maintains an overall factual accuracy of 1.857 on the same large worlds. This gap demonstrates that semantic similarity alone is insufficient when the knowledge structure is hierarchical: a system must understand not just what is relevant, but where it sits in the hierarchy.
 
The scaling behavior of multi-hop queries reinforces this conclusion. While multi-hop proves to be the hardest query type for both systems, the trend tells two very different stories. The baseline's LLM judge score degrades as the world grows (3.450 $\rightarrow$ 2.817 $\rightarrow$ 1.660), consistent with its broader inability to surface structurally relevant nodes at scale. On the other hand, HG-RAG improves with scale (3.233 $\rightarrow$ 3.883 $\rightarrow$ 4.100). Its best multi-hop performance occurs on the largest, most complex world. In small worlds, the traversal advantage is partially offset by context noise from an oversaturated subgraph, and by the baseline facing less retrieval pressure with fewer competing entities. In large worlds, the retrieval is doing exactly what it was designed to do: rounding up a tight, relevant neighborhood from a vast graph. These results suggest that as knowledge graphs grow in complexity, the case for hierarchy-aware retrieval over semantic retrieval only strengthens.

\section*{Limitations and Further Improvements}
The subgraph cap is currently fixed at 15 nodes regardless of world size. In future work, scaling this parameter proportionally with graph size may improve the lackluster performance on small worlds, where the cap captures nearly the entire graph, while preserving the filtering benefit at larger scales.

A significant methodological concern is the evaluation pipeline for open-ended responses. Both the answering model and the grader are the same \texttt{Mistral 7B} instance, which as noted earlier, could possess a self-consistency bias. Compounding this, LLM-based grading carries an inherent hallucination risk independent of model consistency. The grader may reward responses that sound correct rather than being correct. This was observed during testing: the judge awarded a score of 5 to two incorrect responses, likely because they were structured as detailed bullet points and appeared substantive despite being factually wrong. This highlights a fundamental weakness of LLM-as-judge evaluation and suggests that future iterations should use a stronger, separate judge model.

Another pitfall of HG-RAG is the inability to handle queries that do not contain a named entity. To recap, the framework’s first step is to perform one LLM pass to locate a named entity. Without a named-entity of any kind (\texttt{city}, \texttt{country}, \texttt{planet}), there is no way for the system to answer the question. Therefore, an attribute-based query such as \texttt{“Which city has 2 thousand people and exports coal?”} that can be answered using the map, fall outside the current system's capabilities. However, this is not a fundamental flaw in the framework's design. A reverse-lookup or attribute-filtering subroutine could be added to handle these cases. The decision to omit it was deliberate, as the added complexity would have obscured the core contribution of hierarchy-guided retrieval.

\nocite{*}
\bibliography{custom}

\end{document}